\documentclass[]{spie}  

\usepackage{amsmath,amsfonts,amssymb}
\usepackage[colorlinks=true, allcolors=blue]{hyperref}

\usepackage{graphicx}

\title{Unsupervised Segmentation of 3D Medical Images \\ Based on
Clustering and
Deep Representation Learning}

\author[a]{Takayasu Moriya}
\author[a]{Holger R. Roth}
\author[b]{Shota Nakamura}
\author[c]{Hirohisa Oda}
\author[c]{Kai Nagara}
\author[a]{\\Masahiro Oda}
\author[a]{Kensaku Mori}
\affil[a]{Graduate School of Informatics, Nagoya University}
\affil[b]{Nagoya University Graduate School of Medicine}
\affil[c]{Graduate School of Information Science, Nagoya University}

\usepackage{amsmath}
\usepackage{amsfonts}
\usepackage{arydshln} 
\usepackage[subrefformat=parens]{subcaption}

\newcommand{\bvec}[1]{\mbox{\boldmath $#1$}}

\newcommand{\argmin}{\mathop{\rm arg~min}\limits}

\newcommand{\eg}{e.\,g.,\ }

\begin{document}
\maketitle

\begin{abstract}

This paper presents a novel unsupervised segmentation method for 3D medical images. Convolutional neural networks (CNNs) have brought significant advances in image segmentation. However, most of the recent methods rely on supervised learning, which requires large amounts of manually annotated data. Thus, it is challenging for these methods to cope with the growing amount of medical images. This paper proposes a unified approach to unsupervised deep representation learning and clustering for segmentation. Our proposed method consists of two phases. In the first phase, we learn deep feature representations of training patches from a target image using joint unsupervised learning (JULE) that alternately clusters representations generated by a CNN and updates the CNN parameters using cluster labels as supervisory signals. We extend JULE to 3D medical images by utilizing 3D convolutions throughout the CNN architecture. In the second phase, we apply $k$-means to the deep representations from the trained CNN and then project cluster labels to the target image in order to obtain the fully segmented image. We evaluated our methods on three images of lung cancer specimens scanned with micro-computed tomography (micro-CT). The automatic segmentation of pathological regions in micro-CT could further contribute to the pathological examination process. Hence, we aim to automatically divide each image into the regions of invasive carcinoma, noninvasive carcinoma, and normal tissue. Our experiments show the potential abilities of unsupervised deep representation learning for medical image segmentation.

\end{abstract}


\section{Purpose}
\label{sec:intro}  
The purpose of our study is to develop an unsupervised segmentation method of 3D medical images.
Most of the recent segmentation methods using convolutional neural networks (CNNs) rely on supervised learning
that requires large amounts of manually annotated data \cite{long2015fully}.
Therefore, it is challenging for these methods
to cope with medical images due to the difficulty of
obtaining manual annotations.
Thus,
research into unsupervised learning, especially for 3D medical images,
is very promising.
Many previous unsupervised segmentation methods for 3D medical images are based on clustering \cite{garcia2013review}.
However, most unsupervised work in medical imaging was limited to hand-crafted features that were then used with traditional clustering methods to provide segmentation.

In our study,
we investigated whether representations learned by unsupervised deep learning aid in the clustering and segmentation of 3D medical images.
As an unsupervised deep representation learning,
we adopt joint unsupervised learning (JULE) \cite{yang2016joint} based on a framework
that progressively clusters images and learns deep representations
via a CNN.
Our main contribution is to combine JULE with $k$-means \cite{macqueen1967some} for medical image segmentation.
To our knowledge, our methods are the first to employ JULE for unsupervised medical image segmentation.
Moreover, our work is the first to conduct automatic segmentation for pathological diagnosis of micro-CT images.
This work demonstrates that deep representations can be useful for unsupervised medical image segmentation.

There are two reasons why we chose JULE for our proposed method.
The first reason is that JULE is robust against data variation (\eg image type, image size, and sample size) and thus can cope with a dataset composed of 3D patches cropped out of medical images.
Moreover, the range of intensities is different for each medical image.
Thus, we need a learning method that works well with various datasets.
The second reason is that JULE can learn representations that work well with many clustering algorithms.
This advantage allows us to learn representations on a subset of possible patches from a target image and then apply a faster clustering algorithm to the representations of all patches for segmentation.

%
%

\section{METHOD}
The proposed segmentation method has two phases: (1) learning feature representations using JULE
and (2) clustering deep representations for segmentation.
In phase (1),
we conduct JULE in order to learn the representations of image patches
randomly extracted from an unlabeled image.
For use with 3D medical images,
we extend JULE to use 3D convolutions.
The purpose of this phase is to obtain a trained CNN
that can transform image patches to discriminative feature representations.
In phase (2),
we use $k$-means to assign labels to learned representations generated by the trained CNN.

\subsection{Deep Representation Learning}
\begin{figure}[tb]
  \begin{center}
  \includegraphics[keepaspectratio, scale=0.55]{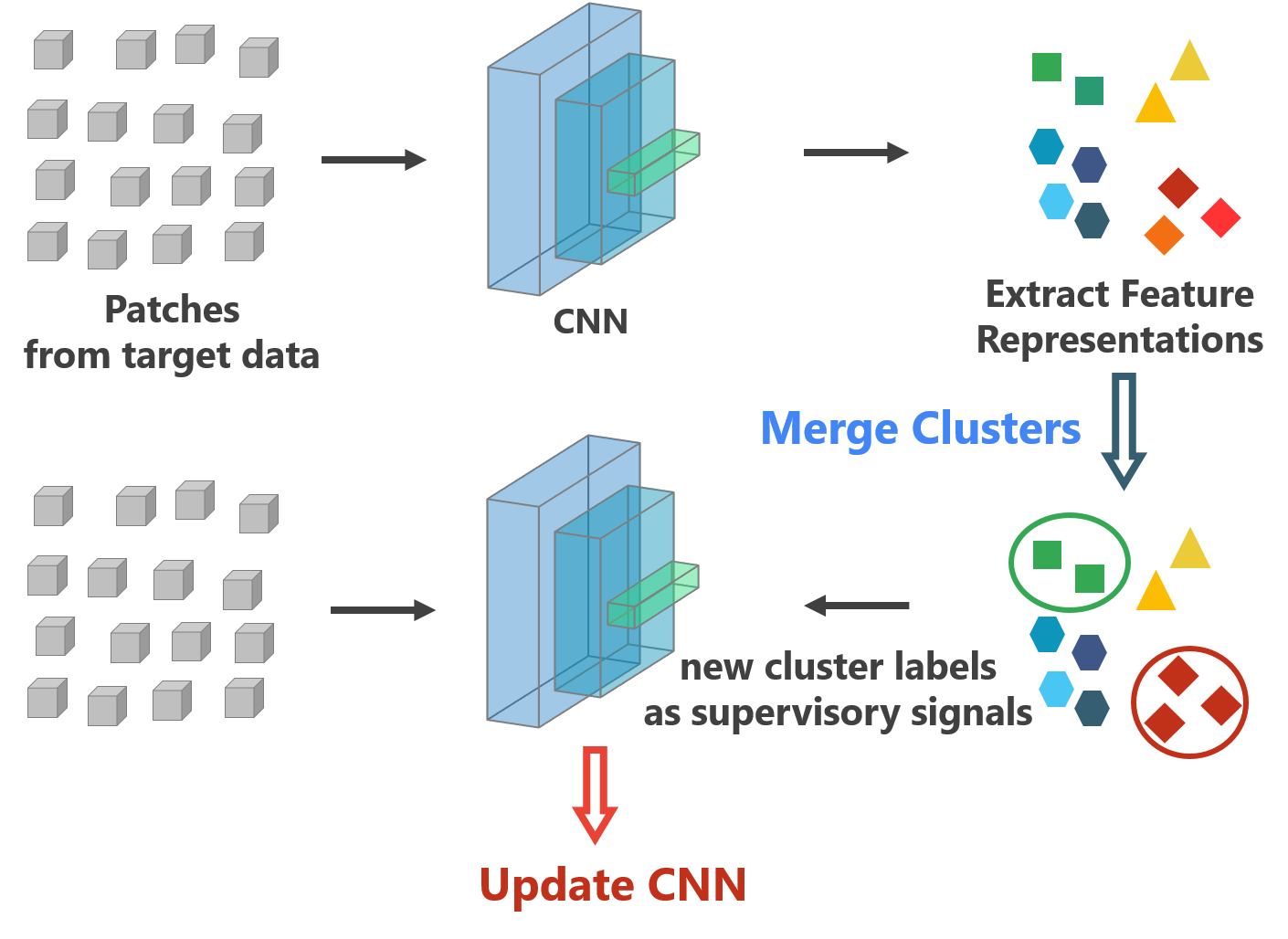}
  \caption{Illustration of a recurrent process at the time of step $t$.
  First, we extract representations $\bvec{X}^{t}$ from training patches $\bvec{I}$
  via a CNN with parameter $\bvec{\theta}^{t}$.
  Next, we merge them and assign new labels $\bvec{y}^{t}$ to $\bvec{X}^{t}$.
  Finally, we input $\bvec{I}$ into the CNN again
  and update the CNN parameters $\bvec{\theta}^{t}$ to $\bvec{\theta}^{t+1}$
  through back propagation from a loss
  calculated using $\bvec{y}^{t}$ as supervisory signals.
  Note that the CNN is initialized with random weights.}
  \label{fig:recurrent}
  \end{center}
\end{figure}

The main idea behind JULE is that
meaningful cluster labels could become supervisory signals for representation learning
and discriminative representations help to obtain meaningful clusters.
Given a set of $n_{s}$ unlabeled image patches $\bvec{I} = \{I_{1},\dots, I_{n_{s}}\}$,
cluster labels for all image patches $\bvec{y} = \{y_{1},\dots, y_{n_{s}}\}$,
and the parameters for representations $\bvec{\theta}$,
the objective function of JULE is formulated as
\begin{equation}
  (\hat{\bvec{y}}, \hat{\bvec{\theta}}) = \argmin_{\bvec{y}, \bvec{\theta}} \mathcal{L}(\bvec{y}, \bvec{\theta}|\bvec{I})
\end{equation}
where $\mathcal{L}$ is a loss function.
JULE tries to find optimal $\hat{\bvec{y}}$ in the forward pass
and optimal $\hat{\bvec{\theta}}$ in the backward pass to minimize $\mathcal{L}$.
By iterating the forward pass and the backward pass,
we can obtain
more discriminative representations and therefore better image clusters.
In the forward pass, we conduct image clustering to merge clusters
using agglomerative clustering
\cite{zhang2012graph}.
In the backward pass, we conduct representation learning
via a 3D CNN
using cluster labels as supervisory signals.
JULE can be interpreted as a recurrent framework
because it iterates merging clusters and learning representations over multiple timesteps
until it obtains the desired number of clusters $C$.
Fig. \ref{fig:recurrent}
shows an overview of a recurrent process at the time of step $t$.

\subsection{Extension to 3D Medical Images}
\label{subsec:extension}
We conducted two extensions of JULE.
One is the extension of the recurrent process
for the CNN training in the backward pass.
Originally,
JULE aims to obtain the final clusters
and finishes when it obtains a desired number of clusters in the forward pass \cite{yang2016joint}.
In contrast, our purpose is to obtain a well-trained CNN.
If we terminate the recurrent process in the final forward pass,
we lose a chance to train the CNN with the final cluster labels.
Therefore, we extended the recurrent process
to train the CNN using the final cluster label in the backward pass.
The intuitive reason is that
the final clusters are the most precise of the entire process
and representations learned with them become more discriminative.
The other is the extension of CNN to support 3D medical images.
Originally,
JULE is a representation learning and clustering method
for 2D images.
We, however,
aim to learn representations using 3D image patches.
Thus,
we extended the CNN architecture of the original JULE \cite{yang2016joint}
to use 3D convolutions throughout the network.

\subsection{Patch Extraction}
Prior to learning representations,
we need to prepare training data composed of small 3D image patches.
These patches are extracted from the unlabeled image,
which is our target for segmentation,
by randomly cropping $n_{s}$ sub-volumes of $w\times w \times w$ voxels.
In many cases of medical image segmentation,
we need to exclude the outside of a scanned object from the training data.
We choose a certain threshold
that can divide the scanned target region from the background
and include only patches whose center voxel intensity is within the threshold.
After extracting training patches,
we centralize them by subtracting out the mean of all intensities and
dividing by the standard deviation,
following Yang et al. \cite{yang2016joint}

\subsection{CNN Architecture}
\begin{figure}[tb]
  \begin{center}
  \includegraphics[keepaspectratio, scale=0.5]{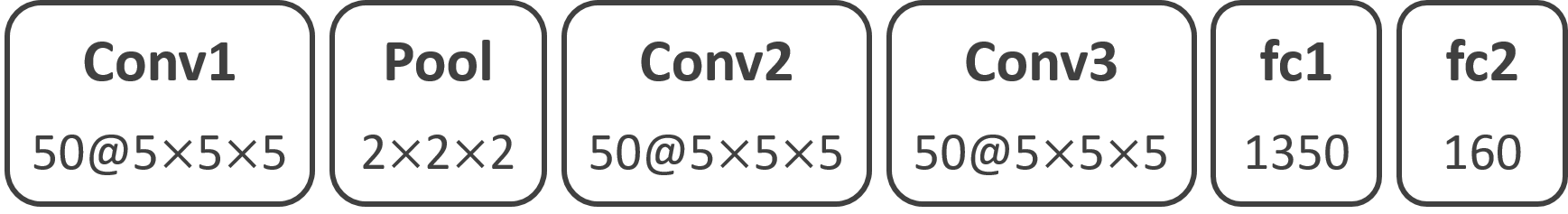}
  \caption{Our CNN architecture has 3 convolutional, 1 max pooling, and 2 fully-connected layers.
  All 3D convolutional kernels are $5\times5\times5$ with stride 1.
  Number of kernels are denoted in each box.
  Pooling kernels are $2\times2\times2$ with stride 2.
  The first fully-connected layer has 1350 neurons,
  and the second one has 160 neurons.}
  \label{fig:img_cnn}
  \end{center}
\end{figure}

Our CNN consists of three convolutional layers,
one max pooling layer, and two fully-connected layers.
The kernels of the second and third convolutional layers are connected
to all kernel maps in the previous layer.
The neurons in the fully-connected layers are connected to
all neurons in the previous layer.
The max pooling layer follows the first convolutional layers.
Batch normalization is applied to the output of each convolutional layer.
A rectified linear unit (ReLU) is used as the nonlinearity after batch normalization.
The second fully-connected layer is followed by the L2-normalization layer.
All of the convolutional layers use 50 kernels of $5\times5\times5$ voxels
with a stride of 1 voxel.
The Max pooling layer has a kernel of $2\times2\times2$ voxels
with a stride of 2 voxels.
The input to the network are image patches of $27\times27\times27$ voxels.
The first fully-connected layer has 1350 neurons
and the second has 160 neurons.
Other parameters for the CNN training, such as learning rate,
are the same as proposed in the original JULE \cite{yang2016joint}.
The CNN architecture is presented in Fig. \ref{fig:img_cnn}.

\subsection{Segmentation}
\begin{figure}[tb]
  \begin{center}
  \includegraphics[keepaspectratio, scale=0.6]{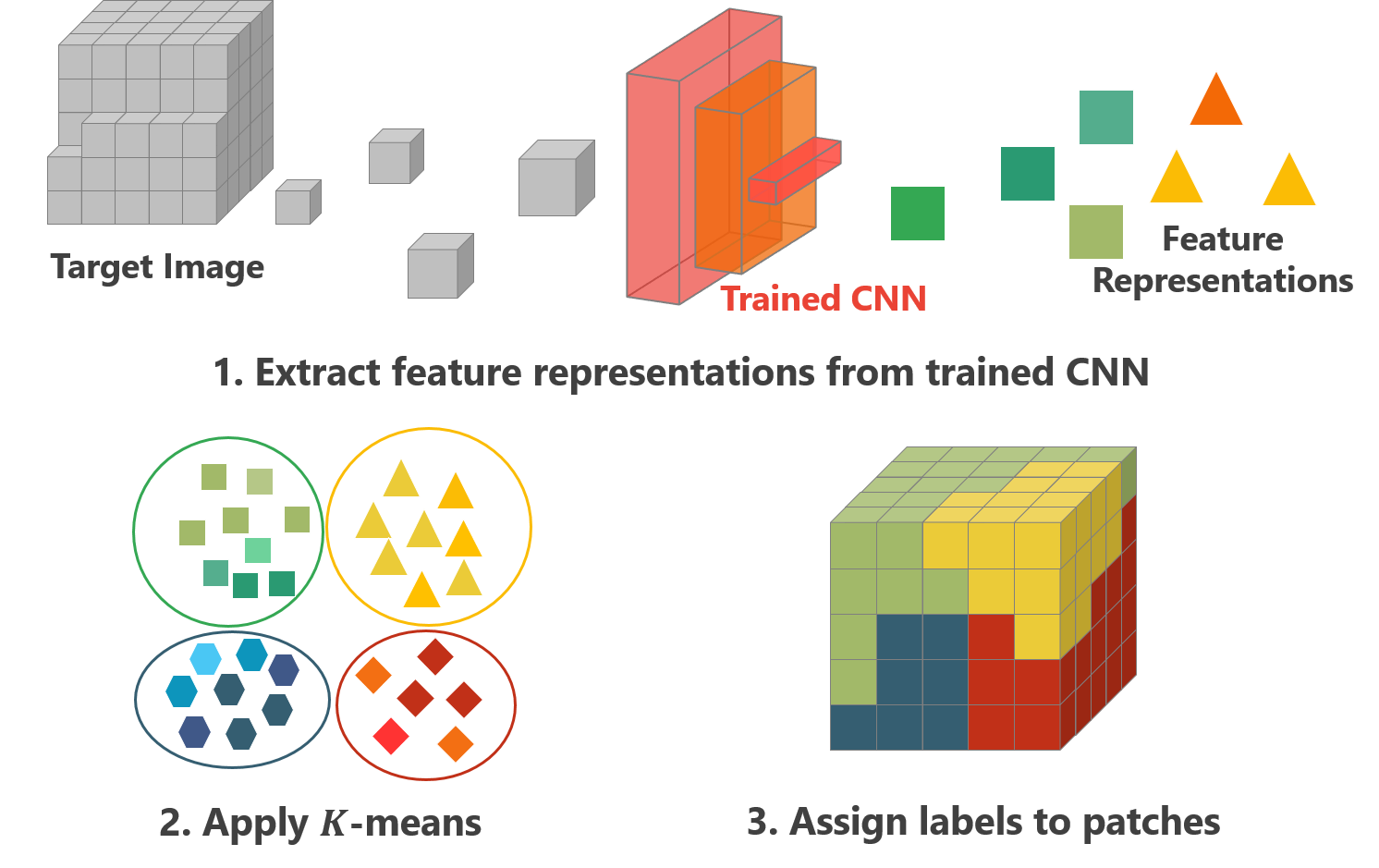}
  \caption{Our segmentation process.
  We first obtain feature representations from a trained CNN
  and then apply conventional $k$-means to them.
  Finally, we assign labels to the patches based on the clustering results.
  (For simplification, we have drawn the figure with a stride equal to $w$.)}
  \label{fig:segmentation}
  \end{center}
\end{figure}

In the segmentation phase,
we first extract a possible number of patches of $w\times w \times w$ voxels from the target image separated by $s$ voxels each.
Note that stride $s$ is not larger than $w$.
As with extracting training patches,
we select only voxels within the scanned sample by thresholding.
The trained CNN transforms each patch into a feature representation.
We then divide the feature representations into $K$ clusters
by $k$-means.
After applying $k$-means,
each representation is assigned a label $l (1 \le l \le K)$
and we need to project these labels onto the original image.
We consider subpatches of $s\times s \times s$ voxels centered in each extracted patch.
Each subpatch is assigned the same label
as the closest cluster representation using Euclidean distance.
This segmentation process is illustrated in Fig. \ref{fig:segmentation}.

\section{EXPERIMENTS AND RESULTS}
\label{sec:results}
\subsection{Datasets}
We chose
three lung cancer specimen images scanned with a micro-CT scanner
(inspeXio SMX-90CT Plus, Shimadzu Corporation, Kyoto, Japan)
to evaluate our proposed method.
The lung cancer specimens from the respective patients
were scanned with similar resolutions.
We aimed to divide each image into three histopathological regions:
(a) invasive carcinoma;
(b) noninvasive carcinoma;
and (c) normal tissue.
We selected these images because
segmenting the regions on micro-CT images based on histopathological features
could contribute to the pathological examination \cite{mori2016macro, nakamura2016micro}.
Detailed information for each image is shown in Table \ref{tab:data}.
\begin{table}[tb]
  \centering
  \caption{Images used in our experiments}
  \label{tab:data}
  \begin{tabular}{c|ccc}
    \hline
    Image & Image Size (voxel) & Resolution ($\mu$m) & Threshold (intensity)\\ \hline
    lung-A & 756$\times$520$\times$545 & 27.1$\times$27.1$\times$27.1 & 4000  \\
    lung-B & 594$\times$602$\times$624 & 29.63$\times$29.63$\times$29.63 & 2820 \\
    lung-C & 477$\times$454$\times$971 & 29.51$\times$29.51$\times$29.51 & 4700  \\
    \hline
  \end{tabular}
\end{table}


\subsection{Parameter Settings}
For JULE, we randomly extracted 10,000 patches of $27\times27\times27$ voxels  from a target image.
We set the number of final clusters $C$ to 100 for lung-A and lung-C,
to 10 for lung-B,
which are the stopping conditions of agglomerative clustering.
Other parameters are the same as in the original JULE \cite{yang2016joint}.
After representation learning,
we extracted patches of $27\times27\times27$ voxels with a stride of five voxels
and processed them by the trained CNN to obtain a 160 dimensional representation for each patch.
For segmentation,
we applied the conventional $k$-means to the feature representations,
setting $K$ to 3.

\subsection{Evaluations}
\begin{figure}[tb]
  \begin{tabular}{cccc}
    \begin{minipage}[t]{0.5\linewidth}
     \centering
     \includegraphics[keepaspectratio, scale=0.24]
     {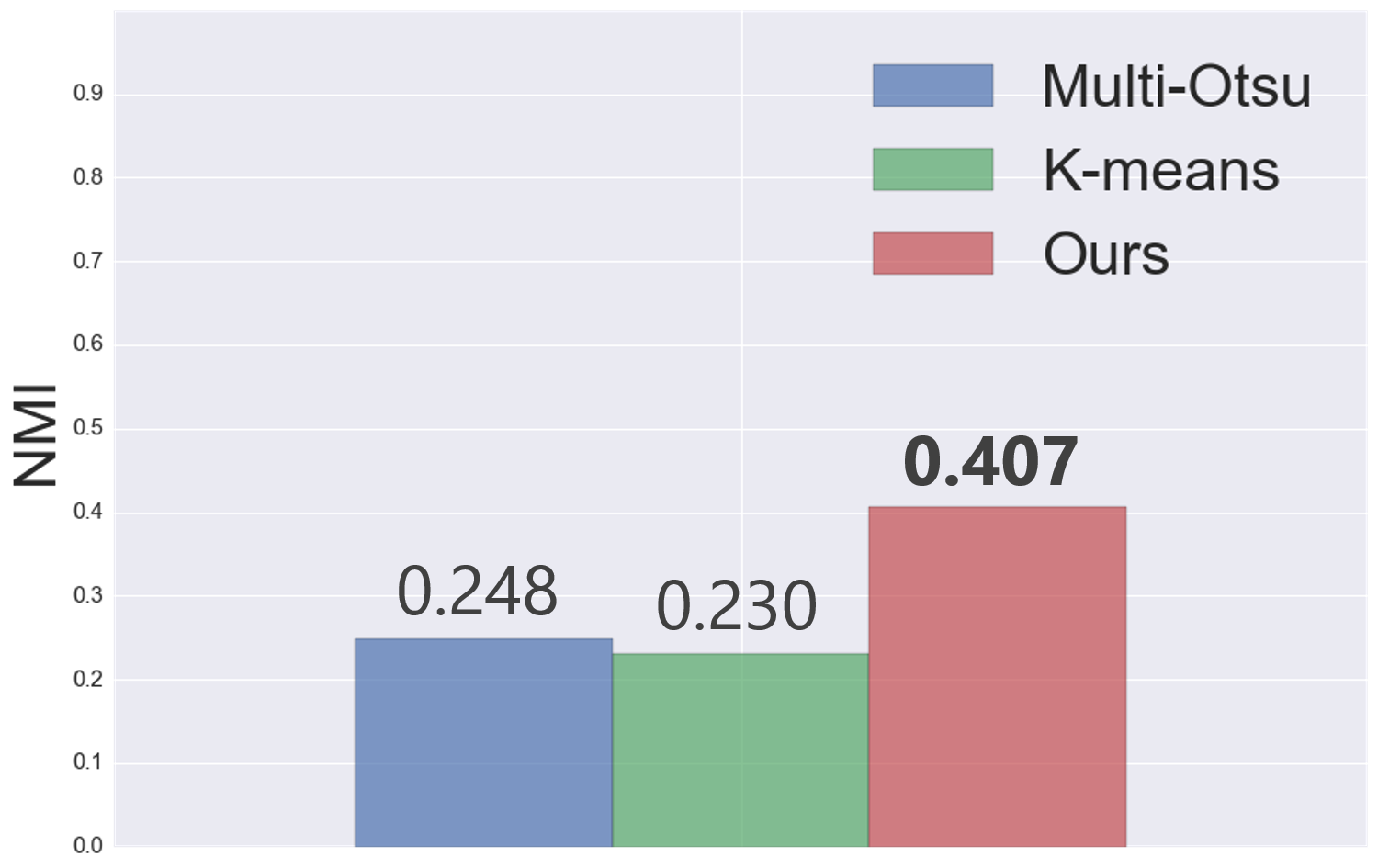}
   	\subcaption{NMI comparison on lung-A}
   	\label{subfig:bar1}
    \end{minipage}
    \begin{minipage}[t]{0.5\linewidth}
     \centering
     \includegraphics[keepaspectratio, scale=0.24]
   	{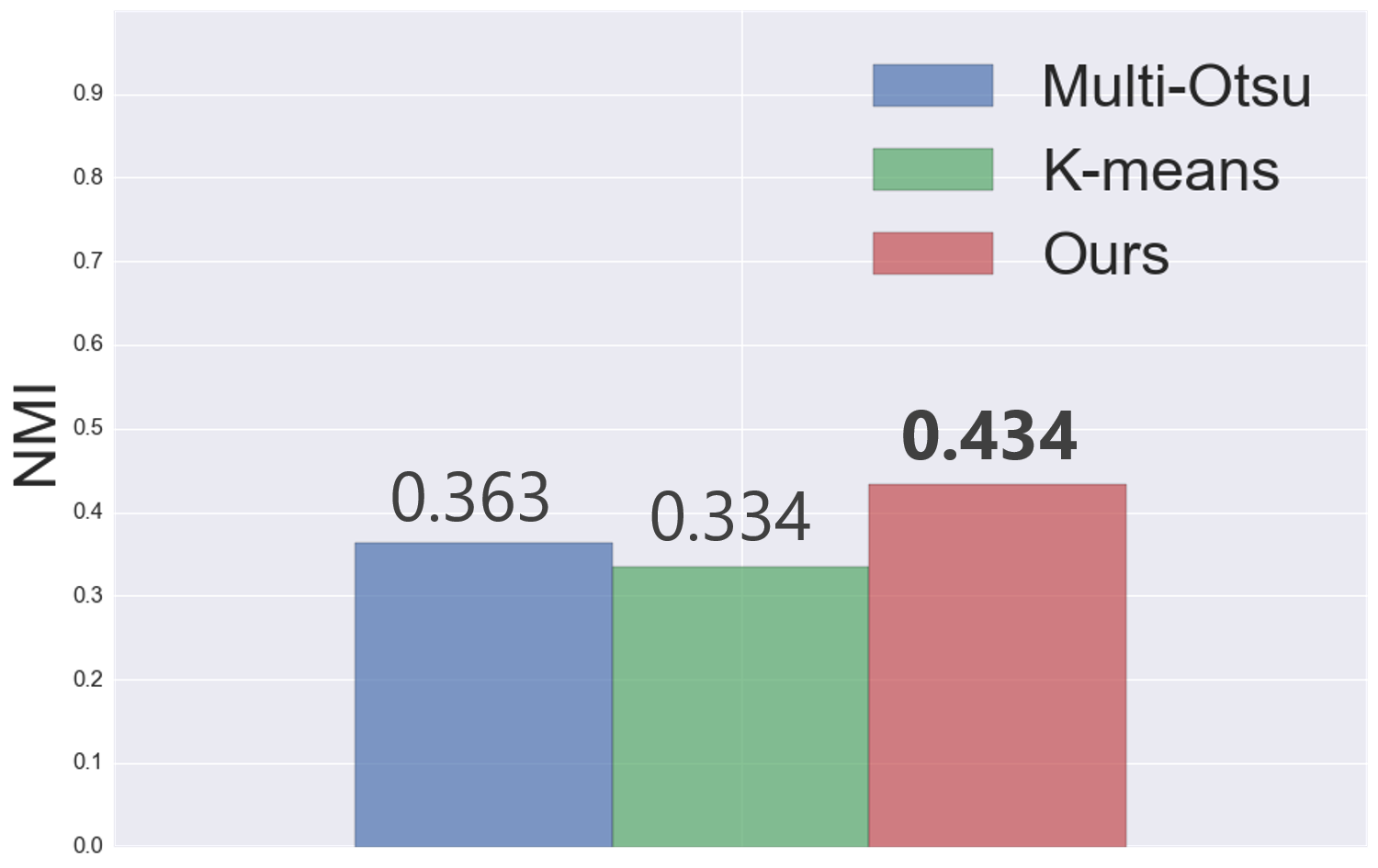}
   	\subcaption{NMI comparison on lung-B}
   	\label{subfig:bar2}
    \end{minipage}\\
    \begin{minipage}[t]{0.5\linewidth}
     \centering
     \includegraphics[keepaspectratio, scale=0.24]
   	{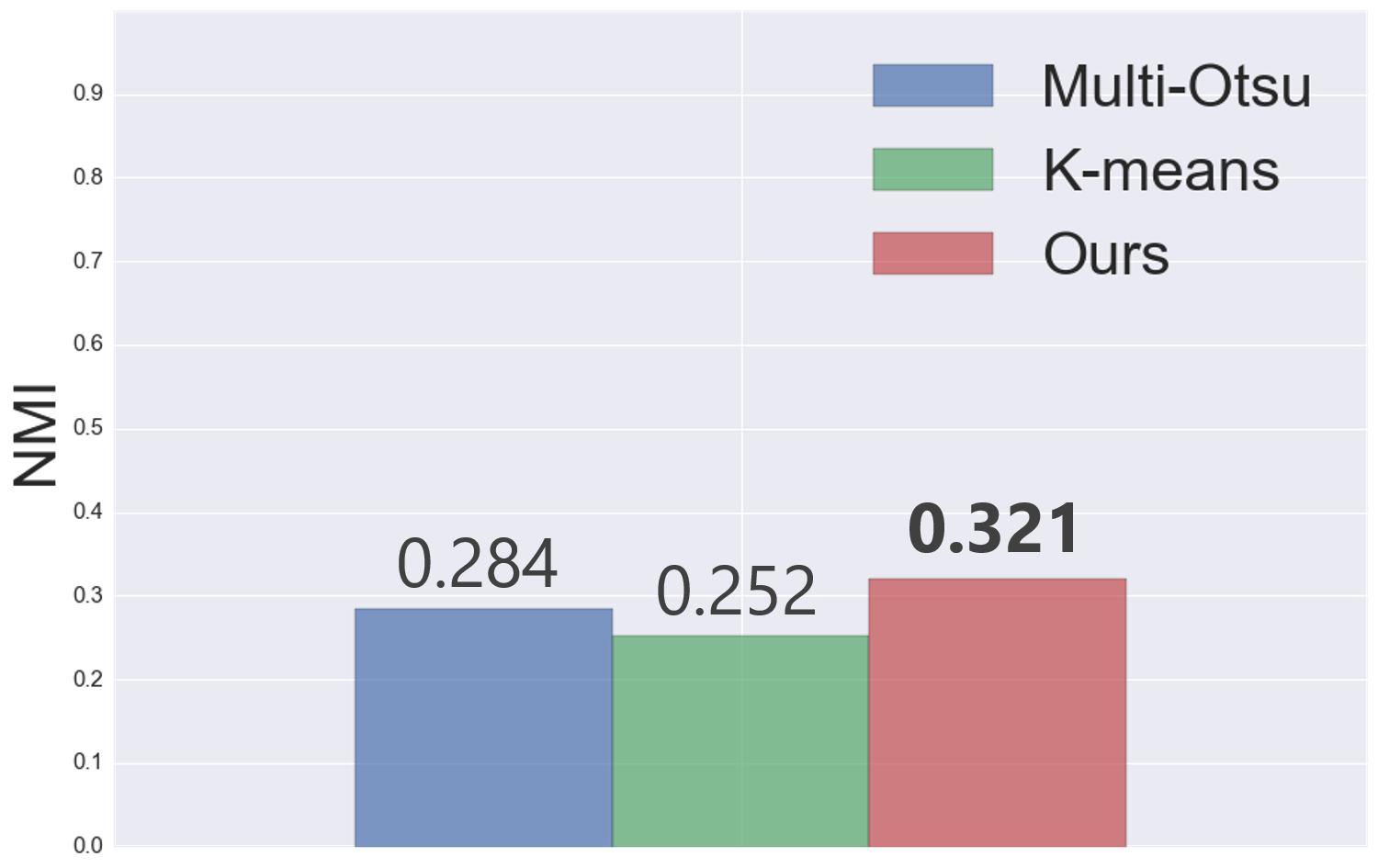}
   	\subcaption{NMI comparison on lung-C}
   	\label{subfig:bar3}
    \end{minipage}
    \begin{minipage}[t]{0.5\linewidth}
     \centering
     \includegraphics[keepaspectratio, scale=0.24]
   	{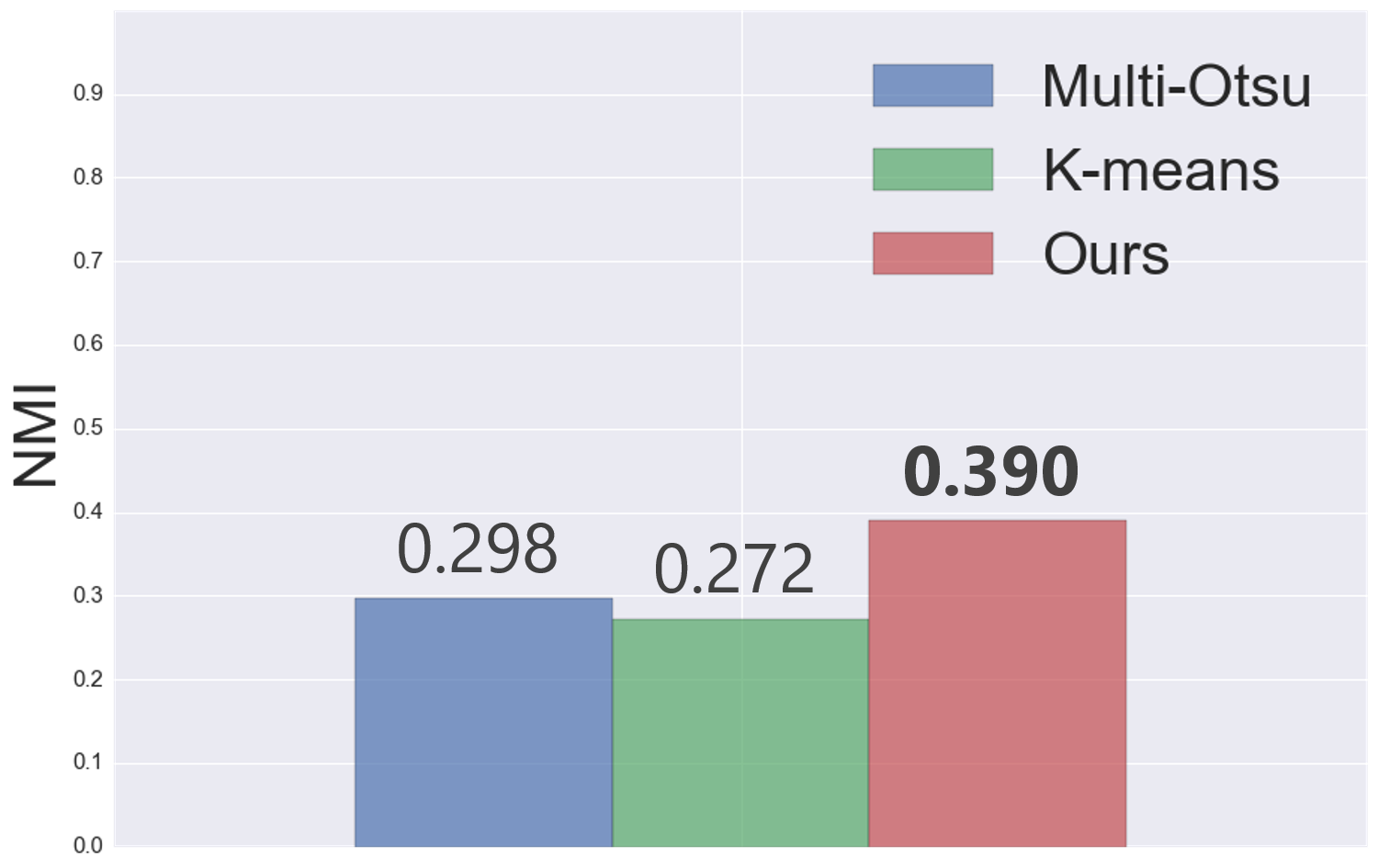}
   	\subcaption{Average NMI comparison}
   	\label{subfig:bar4}
    \end{minipage}
  \end{tabular}
 \caption{NMI comparison on three datasets.
 Our method outperformed traditional unsupervised methods.
 }
 \label{fig:nmi}
\end{figure}

\begin{figure}[tb]
  \begin{center}
  \includegraphics[keepaspectratio, scale=0.8]{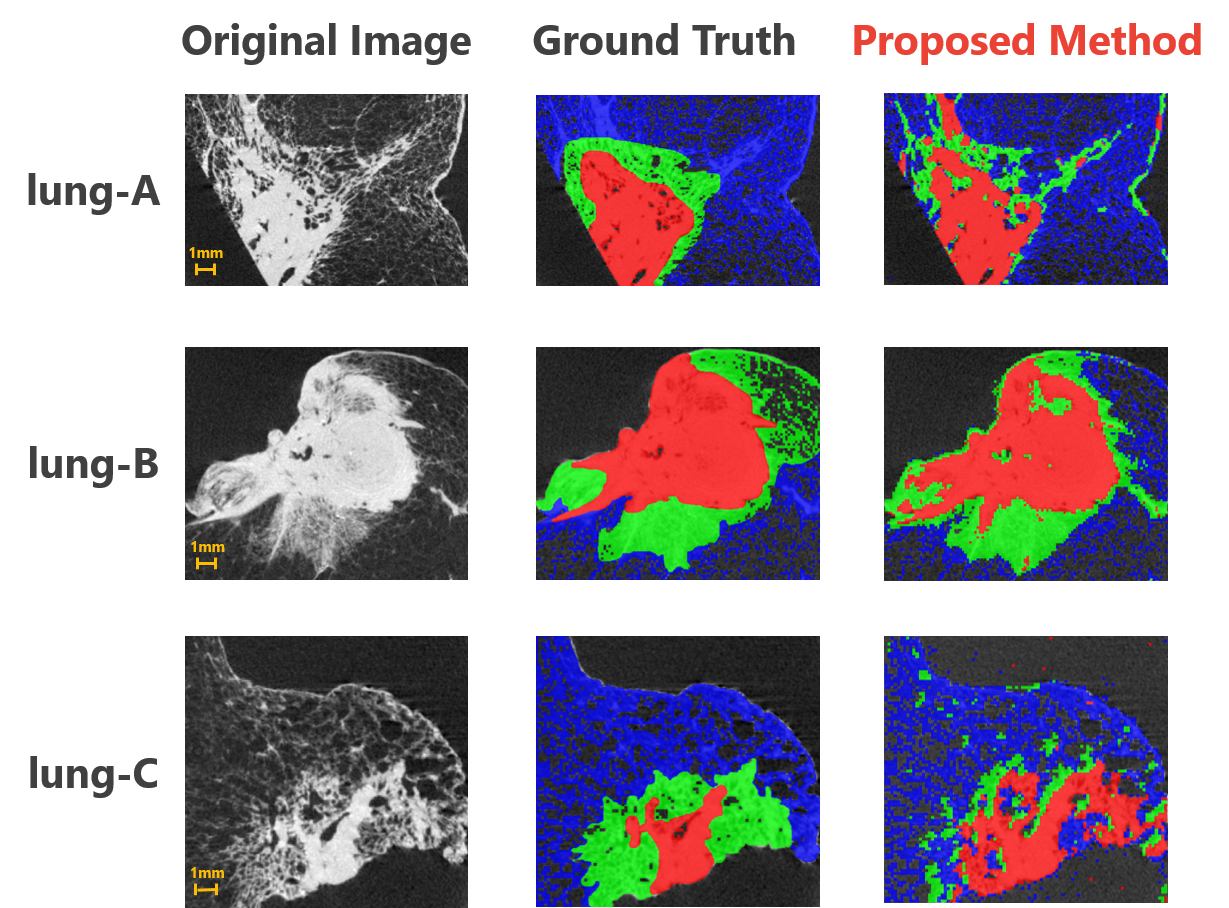}
  \caption{Segmentation results of lung-A (top line), lung-B (middle line),
  and lung-C (bottom line).
  In the ground truth,
  the red, green, and blue regions correspond to the region of
  invasive carcinoma, noninvasive carcinoma, and normal tissue, respectively.
  In the results of the JULE-based segmentation,
  colors indicate the same cluster, but are assigned at random.}
  \label{fig:result_images}
  \end{center}
\end{figure}

We used normalized mutual information (NMI) \cite{strehl2002cluster} to measure segmentation accuracy.
A larger NMI value means more precise segmentation results.
We used seven manually annotated slices for evaluation.
We compared the proposed method with traditional $k$-means segmentation and multithreshold Otsu method \cite{otsu1979threshold}.
We also evaluated the average NMI of each method across the datasets.
The results are shown in
Fig. \ref{fig:nmi}.
In each figure, the best performance NMI for each $K$ is in bold.
As shown in all of the figures,
JULE-based segmentation outperformed traditional unsupervised methods.
While the NMI scores of our methods are not high, qualitative evaluation shows promising results of our proposed method (see Fig. \ref{fig:result_images}).
The qualitative examples show that JULE-based segmentation divided normal tissue region from the cancer region,
including invasive carcinoma and noninvasive carcinoma, well.


\section{Discussions}
Qualitative evaluations demonstrate that
JULE can learn features that divide higher intensity regions
from lower intensity regions.
This is because, generally,
regions of invasive and noninvasive carcinoma
have substantially high intensities,
whereas normal tissue regions have low intensities.
%
Moreover, for lung-A and lung-B,
JULE divided invasive carcinoma from noninvasive carcinoma.
This results shows the potential ability to learn features
that reflect variation in intensity.
This is because, seemingly, invasive carcinoma and normal tissue typically have a small variation of intensities,
whereas noninvasive carcinoma has a large variation of intensities.

\section{CONCLUSION}
We proposed an unsupervised segmentation using JULE that alternately learns deep representations and image clusters.
We demonstrated the potential abilities of unsupervised medical image segmentation using deep representations.
Our segmentation method could be applicable to many other applications
in medical imaging.

\section*{ACKNOWLEDGMENTS}
This research was supported by the Kakenhi by MEXT and JSPS (26108006, 17K20099) and the JSPS Bilateral International Collaboration Grants.
%
%
%

\bibliography{report} 

\begin{thebibliography}{1}

\bibitem{long2015fully}
Long, J., Shelhamer, E., and Darrell, T., ``Fully convolutional networks for
  semantic segmentation,'' in [{\em IEEE CVPR}{\nolinebreak\hspace{0.1em}]},
  3431--3440 (2015).

\bibitem{garcia2013review}
Garc{\'\i}a-Lorenzo, D., Francis, S., Narayanan, S., Arnold, D.~L., and
  Collins, D.~L., ``Review of automatic segmentation methods of multiple
  sclerosis white matter lesions on conventional magnetic resonance imaging,''
  {\em Medical Image Analysis}~{\bf 17},  1--18 (2013).

\bibitem{yang2016joint}
Yang, J., Parikh, D., and Batra, D., ``Joint unsupervised learning of deep
  representations and image clusters,'' in [{\em IEEE
  CVPR}{\nolinebreak\hspace{0.1em}]},   5147--5156 (2016).

\bibitem{macqueen1967some}
MacQueen, J. et~al., ``Some methods for classification and analysis of
  multivariate observations,'' in [{\em Proceedings of the fifth Berkeley
  Symposium on Mathematical Statistics and
  Probability}{\nolinebreak\hspace{0.1em}]},   {\bf 1},  281--297 (1967).

\bibitem{zhang2012graph}
Zhang, W., Wang, X., Zhao, D., and Tang, X., ``Graph degree linkage:
  Agglomerative clustering on a directed graph,'' in [{\em
  ECCV}{\nolinebreak\hspace{0.1em}]},   {\bf 7572},  428--441 (2012).

\bibitem{mori2016macro}
Mori, K., ``From macro-scale to micro-scale computational anatomy: a
  perspective on the next 20 years,'' {\em Medical Image Analysis}~{\bf 33},
  159--164 (2016).

\bibitem{nakamura2016micro}
Nakamura, S., Mori, K., Okasaka, T., Kawaguchi, K., Fukui, T., Fukumoto, K.,
  and Yokoi, K., ``Micro-computed tomography of the lung: Imaging of alveolar
  duct and alveolus in human lung,'' in [{\em D55. LAB METHODOLOGY AND
  BIOENGINEERING: JUST DO IT}{\nolinebreak\hspace{0.1em}]},   A7411--A7411,
  American Thoracic Society (2016).

\bibitem{strehl2002cluster}
Strehl, A. and Ghosh, J., ``Cluster ensembles---a knowledge reuse framework for
  combining multiple partitions,'' {\em Journal of machine learning
  research}~{\bf 3}(Dec),  583--617 (2002).

\bibitem{otsu1979threshold}
Otsu, N., ``A threshold selection method from gray-level histograms,'' {\em
  IEEE transactions on systems, man, and cybernetics}~{\bf 9}(1),  62--66
  (1979).

\end{thebibliography}
\bibliographystyle{spiebib} 

\end{document}